# A Dynamic Fuzzy Rule and Attribute Management Framework for Fuzzy Inference Systems in High-Dimensional Data

KE LIU, JING MA, and EDMUND M-K LAI, Department of Data Science and Artificial Intelligence, Auckland University of Technology, New Zealand

This paper presents an Adaptive Dynamic Attribute and Rule (ADAR) framework designed to address the challenges posed by high-dimensional data in neuro-fuzzy inference systems. By integrating dual weighting mechanisms—assigning adaptive importance to both attributes and rules—together with automated growth and pruning strategies, ADAR adaptively streamlines complex fuzzy models without sacrificing performance or interpretability. Experimental evaluations on four diverse datasets – Auto MPG (7 variables), Beijing PM2.5 (10 variables), Boston Housing (13 variables), and Appliances Energy Consumption (27 variables), show that ADAR-based models achieve consistently lower Root Mean Square Error (RMSE) compared to state-of-the-art baselines. On the Beijing PM2.5 dataset, for instance, ADAR-SOFENN attained an RMSE of 56.87 with nine rules, surpassing traditional ANFIS [12] and SOFENN [16] models. Similarly, on the high-dimensional Appliances Energy dataset, ADAR-ANFIS reached an RMSE of 83.25 with nine rules, outperforming established fuzzy logic approaches and interpretability-focused methods such as APLR. Ablation studies further reveal that combining rule-level and attribute-level weight assignment significantly reduces model overlap while preserving essential features, thereby enhancing explainability. These results highlight ADAR's effectiveness in dynamically balancing rule complexity and feature importance, paving the way for scalable, high-accuracy, and transparent neuro-fuzzy systems applicable to a range of real-world scenarios.

## 1 Introduction

Neuro-fuzzy Systems (NFS), first proposed in the 1990s, seek to combine the interpretability of fuzzy IF-THEN rules with the learning capabilities of artificial neural networks [26]. Different types of NFS have beem successfully applied to areas such as industrial process control [2, 18], medical pattern recognition [7, 9, 24], and financial forecasting [15, 30]. However, like many other machine learning systems, NFS suffers from the "curse of dimensionality" as the number of input features increases. Attempts have been made to address this shortcoming by using manually defined rules or fixed feature subsets [32]. But these methods typically lack adaptive mechanisms to handle data heterogeneity in real-world applications [35].

Authors' Contact Information: Ke Liu, ke.liu@autuni.ac.nz; Jing Ma, jing.ma@aut.ac.nz; Edmund M-K Lai, edmund.lai@aut.ac.nz, Department of Data Science and Artificial Intelligence, Auckland University of Technology, Auckland, New Zealand.

Despite notable advancements in tackling high-dimensional data, several key challenges remain. Firstly, techniques such as autoencoders [11], particle swarm optimization (PSO) [13], and modular NFS [12] designs have improved robustness and computational efficiency. But they often operate without a unified framework that adaptively integrates feature selection, rule management, and performance optimization. This leads to fragmented, domain-specific solutions that are difficult to generalize across heterogeneous datasets or dynamically changing environments.

Secondly, importance-weighting mechanisms have been employed in modelslike Fuzzy-ViT [17] and GFAT [8] to put more emphasis on critical input features. However, such models remain largely restricted to specific applications such as medical imaging, cancer metastasis prediction, and ocean wave height forecasting [1, 8, 17]. Furthermore, they do not have explicit mechanisms for dynamic rule and attribute management, limiting transparency and interpretability when dealing with high-dimensional and evolving data. As a result, these weight-based methods struggle to provide a scalable solution that balances predictive accuracy and model explainability across diverse domains.

Thirdly, current strategies for handling high-dimensional data lack systematic mechanisms to regulate rule complexity and feature importance cohesively. Many approaches still depend on manually defined rules or simplistic pruning algorithms, which are insufficient for capturing the complexities of real-world data distributions that evolve over time and exhibit latent interactions among multiple attributes [21, 31, 33]. This not only undermines model interpretability but also adversely affects performance in domains where data characteristics and operational conditions are subject to rapid change.

In this paper, we propose the Adaptive Dynamic Attribute and Rule (ADAR) framework to address the challenges mentioned above.It incorporates dual importance-weighting mechanisms that operate on both attributes and rules, in contrast to existing schemes that treat them separately. In this way, ADAR is able to regulate rule complexity and feature importance under a single cohesive framework. This ensures scalability and transparency across diverse datasets, filling a critical gap in neuro-fuzzy research. To the best of our knowledge, this is the first time adaptive rule and feature weighting, dynamic structural regulation, and neuro-fuzzy logic are integrated into a single framework. ADAR has the following key characteristics:

(1) **Dual Weighting Mechanism**: By assigning importance weights to both attributes and rules, the proposed framework is able to achieve precise control of the model complexity, improving both performance and interpretability.
(2) **Adaptive Management of Rules and Attributes**: Automated growth and pruning strategies enable the model to adapt the number of rules and attributes according to the complexity of the dataset, overcoming the limitations of fixed or manually defined structures.
(3) **Addressing Challenges of High-Dimensional Data**: Attribute pruning and weight-guided learning effectively mitigate the "curse of dimensionality", allowing fuzzy logic systems to be applied to complex, high-dimensional environments.
(4) **Improved Model Explainability**: By explicitly highlighting the importance of each attribute and rule, the ADAR framework enhances transparency in the decision-making process, thereby contributing to the advancement of explainable artificial intelligence.

## 2 Managing High-dimensional Data in NFS

Managing the exponential growth of model parameters has been a concern when NFS is applied to large datasets. Early efforts primarily focused on extending traditional fuzzy logic architectures to handle increased dimensionality. But these approaches often resulted in overly complex models that were difficult to train and interpret [5]. As datasets grew in size and diversity, researchers began incorporating dimensionality reduction methods such as principal component analysis (PCA) and autoencoders, to project high-dimensional inputs into lower dimensional subspaces [34]. For instance, Pirmoradi et al. [19] proposed a self-organizing deep neuro-fuzzy inference system that addresses the challenges of high-dimensional miRNA genomic data in kidney cancer subtype classification. This method integrates a correlation-based filter for feature selection, leveraging the Arithmetic-Geometric Mean dispersion measure, and employs a deep auto-encoder to compress the high-dimensional data used as input to the NFS. This approach significantly reduces computational complexity while maintaining robust predictive accuracy. Similarly, R.K. Sevakula and N.K. Verma [23] utilized a deep auto-encoder for dimensionality reduction. The output of the deep auto-encoder acts as a compressed representation of the input data for an NFS. Afshin Shoeibi et al. [27] combined autoencoder and PCA as dimensionality reduction techniques for an Adaptive Neuro-Fuzzy Inference System (ANFIS) to predict epileptic seizures. Although these preprocessing techniques help alleviate computational demands, they often separate the transformation process from the fuzzy inference layer, leading to loss of valuable information which is critical for effective rule generation.

Recent research has increasingly emphasized the integration of optimization algorithms with neuro-fuzzy models to improve performance in high-dimensional tasks. A notable example is the Particle Swarm Optimization (PSO)-based ANFIS model introduced by Shahaboddin et al. [25]. This model utilizes the PSO algorithm to optimize the parameters of membership functions in fuzzy systems, achieving improved robustness and accuracy in modeling complex nonlinear problems. At the same time, it is also able to reduces model complexity.

An alternative approach makes use of hierarchical and modular NFS frameworks. The high-dimensional input space is divided into smaller, manageable subspaces, where localized fuzzy inference is conducted independently before combining the results into a final prediction. Compared to traditional global modelling, these strategies offer distinct advantages in interpretability and performance optimization. Siminski et al. [28] expanded on this concept by integrating fuzzy biclustering, which clusters objects and features simultaneously to create subspace-specific fuzzy rules. In this way, features that are of high importance in particular dimensions are emphasized.

Another promising direction focuses on integrating efficient feature processing techniques to enhance adaptability in high-dimensional environments. For example, Sajid et al. [22] introduced the NF-RVFL model, which combines Random Vector Functional Link Networks (RVFL) with NFS. It employs clustering methods such as K-means and fuzzy C-means to provide initial values for the fuzzy set centers and utilizes random projection alongside direct connection strategies. By making use of redundancy in high-dimensional inputs while preserving critical original feature information, the model enhances feature diversity representation and training efficiency. This innovation strikes a balance between computational complexity and model performance, even under complex data distributions.

While the strategies described above have significantly enhanced NFS performance in high-dimensional scenarios, integrating these approaches into a flexible and scalable unified framework remains a challenge. They generally focus on localized optimization and dimensionality reduction techniques, lacking a way to effectively integrate hierarchical modelling, feature selection, and optimization algorithms. This often results in fragmented models with limited adaptability to heterogeneous or dynamic environments. In addition, as application contexts continue to diversify, there is a growing need for models capable of managing cross-subspace rule interactions and supporting real-time updates. Current methodologies struggle to achieve a dynamic balance between performance, interpretability, and scalability in a systematic way.

## 3    The ADAR Framework

We propose the Adaptive Dynamic Attribute and Rule (ADAR) framework to overcome the short-comings of current NFS mentioned above.It aims to enhance both the performance and the inter-pretability of existing NFS such as ANFIS [12]. Learnable weighting mechanism for both attributes and rules are introduced into the training process of the NFS. At the same time, a rule growing and pruning process as well as an attribute pruning process are also introduced. Thus, the three primary elements in this framework are: (1) initialization, (2) attribute weighting and pruning, and (3) rule growing and pruning. Figure 1 illustrates the ADAR framework integrated within the ANFIS structure. It presents the architectural flow from input to output, highlighting how the Attribute Weighting Mechanism and Rule Weighting Mechanism are embedded to facilitate interpretable and efficient model behavior.

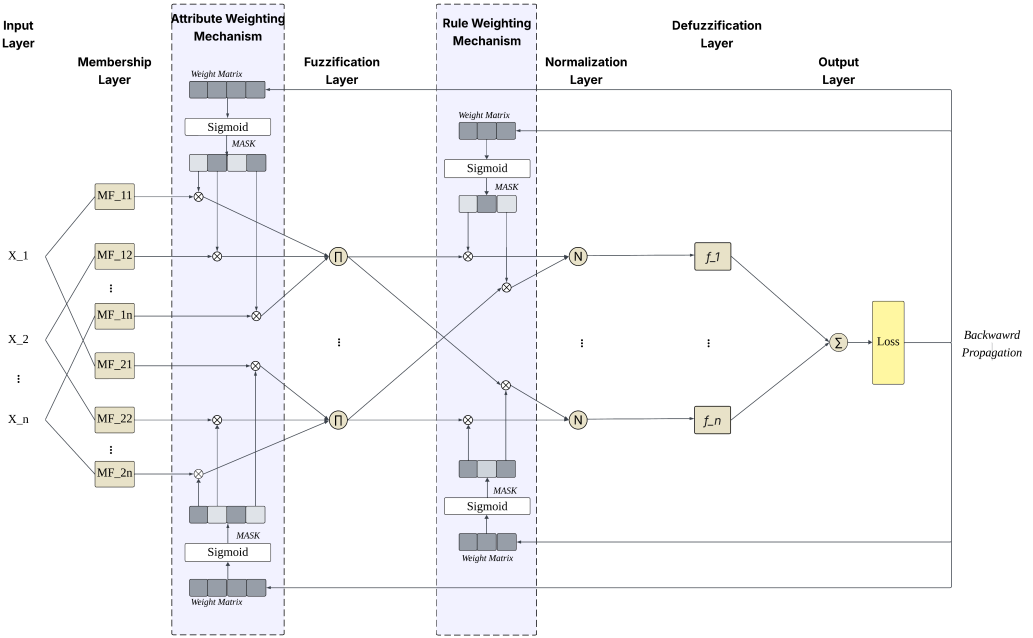

Fig. 1.  Architecture of the ADAR-ANFIS framework

### 3.1    Initialization

Initialization involves determining the initial number and parameters of the fuzzy rules as well as the importance weights for both the data attributes and the fuzzy rules. The normalized training data are first partitioned into clusters through K-means clustering. The centroid of each cluster is then used as the initial value of the means of the Gaussian membership functions for each rule. The shape and support of the rules which are quantified by the standard deviations of the Gaussian functions, are initialized by the per-feature standard deviation of the data points in each cluster. This allows the width of each membership function to reflect the local data dispersion, providing a data-driven, rule-specific foundation for the fuzzy partitions. The initial attribute importance weights are randomly sampled from a Gaussian distribution, introducing initial variability across dimensions. The rule weights are typically initialized to one, indicating equal contribution from all rules at the start.

## 3.2 Neural-fuzzy System with Weighting Mechanism

During training, both types of importance weights are treated as fully differentiable parameters and updated via backpropagation. The model further incorporates L1 regularization on these weights to promote sparsity—automatically suppressing irrelevant input features and redundant rules. This strategy enhances the interpretability of the model by selectively amplifying informative patterns, while simultaneously enabling automatic structural pruning during training. As a result, the model evolves into a more compact, efficient, and semantically meaningful fuzzy inference system.

During training iterations, the Attribute Pruning (AP) mechanism is periodically activated. AP leverages attribute importance weights to identify and remove attributes with consistently low contributions, thus simplifying the model structure and enhancing interpretability. Following pruning, performance variations are evaluated to determine if pruning significantly degrades performance. If that is the case, then the previous structure is restored to maintain balance between complexity and accuracy.

ADAR also employs a Rule Growing and Pruning (RG&P) strategy to periodically adjust the size of the rule base. Rule pruning removes redundant rules by assessing their importance weights and eliminating those with persistent low influence. Conversely, rule growing identifies poorly represented regions with high prediction errors, prompting the creation of new rules to capture these complex data patterns effectively. Whenever structural modifications occur, the optimizer is reinitialized to align parameter optimization with the current model architecture. These processes are described in more detail in the following sections.

## 3.3 Attribute Weighting Mechanism

Under the ADAR framework, a learnable importance weight $\alpha_{l,i}$ is associated with each attribute $i$ within every rule $l$.

These weights reflect the relative importance of each feature within the associated fuzzy rule. By learning such weights during training, the model effectively filters the input feature space, focussing on attributes that significantly impact the predictive task while ignoring those with minimal contribution.

A trainable parameter matrix $\mathbf{W}_a \in \mathbb{R}^{L \times D}$

was introduced, where $L$ is the number of rules and $D$ the number of input attributes. This matrix captures the attribute weight logits, with each element $\alpha_{l,i}$ representing the raw (pre-activation) importance score of feature $i$ under rule $l$. These logits are transformed using a sigmoid activation function to produce the final importance weights, which are further modulated by a binary attribute mask that enforces sparsity at the rule level.

The importance weight $\alpha_{l,i}$ is computed as follows:

$$\alpha_{l,i} = \sigma(w_{a,l,i}) \cdot m_{l,i} \tag{1}$$

Here, $\sigma(\cdot)$ represents the sigmoid function, ensuring that $\alpha_{l,i}$ remains within the range $[0, 1]$. The variable $m_{l,i}$ serves as an attribute mask, initialized to 1 for all attributes and updated during the pruning process. When an attribute is pruned from a rule, its mask is set to 0, guaranteeing that the attribute no longer affects the fuzzy inference process.

The term $w_{a,l,i}$ is a trainable parameter that directly governs the importance weight before activation. It is learned end-to-end via gradient descent, and its magnitude determines the influence of feature $i$ in rule $l$. A large positive value of $w_{a,l,i}$ leads to $\alpha_{l,i}$ approaching 1, indicating high relevance, whereas a large negative value pushes $\alpha_{l,i}$ toward 0, signifying low importance.

## 3.4 Attribute Pruning (AP)

High-dimensional feature spaces often include attributes that provide minimal contribution to the predictive task, even with the Attribute Weighting Mechanism in place. While importance weights help identify the importance of each attribute within a rule, some attributes may consistently remain underutilized. The main objective of AP is to remove attributes that, despite being initially included, consistently fail to demonstrate significant importance. After the Attribute Weighting Mechanism undergoes sufficient training to adjust importance weights, attributes with consistently low weights values are pruned from the corresponding rules. This process reduces the effective dimensionality of the input space for those rules, mitigates overfitting, and enhances computational efficiency.

Let $\alpha_{l,i}$ represent the importance weight for attribute $i$ in rule $l$, as defined by the Attribute Weight Mechanism. If an attribute $i$ in rule $l$ consistently maintains an importance weight below a predefined pruning threshold $\theta_{\text{attr}}$:

$$\alpha_{l,i} < \theta_{\text{attr}} \tag{2}$$

for a prolonged duration or after a specified number of training epochs, the corresponding attribute mask $m_{l,i}$ is set to zero, effectively excluding attribute $i$ from the inference process in rule $l$:

$$m_{l,i} = 0 \quad \text{if} \quad \alpha_{l,i} < \theta_{\text{attr}}. \tag{3}$$

Here, $\theta_{\text{attr}}$ is a small hyperparameter that defines the minimum acceptable level of attribute relevance. This criterion ensures that only attributes with sufficient importance remain active in the model.

## 3.5 Rule Weighting Mechanism

As the complexity of fuzzy neural networks growths, managing a large number of rules becomes increasingly challenge. While Attribute Weighting Mechanism prioritizes the most relevant features within each rule, the Rule Weighting Mechanism is designed to dynamically evaluate and emphasize the significance of entire rules. By assigning and learning a distinct importance weight for each rule, this mechanism directs the model to focus on rules that significantly contribute to the predictions while de-emphasizing or eliminating those that provide limited or redundant information.

The Rule Weighting Mechanism introduces a learnable parameter $\beta_l$ to each rule $l$. Intuitively, $\beta_l$ represents the overall importance of rule $l$ in the inference process. Rules that consistently fail to reduce residual errors will maintain low importance weights, indicating their minimal contributes to the model's decision-making. Conversely, rules that capture meaning patterns or enhance predictions will achieve higher importance weights, ensuring their impact on the final output is appropriately amplified.

For each rule $l$, a learnable parameter $w_{r,l}$ is introduced. To ensure that the rule importance weight $\beta_l$ remains within the range $[0, 1]$, a sigmoid activation function is applied:

$$\beta_l = \sigma(w_{r,l}) \tag{4}$$

Here, $\sigma(\cdot)$ represents the sigmoid function. A higher value of $\beta_l$ denotes a more influential rule, whereas a lower $\beta_l$ indicates diminished significance. During training, $w_{r,l}$ is updated through gradient-based optimization, allowing the model to dynamically adjust the importance of each rule based on the data.

### 3.6 Rule Growing and Pruning (RG&RP)

While the Attribute Weighting Mechanism and Attribute Pruning (AP) steps refine the input space within each rule, the overall complexity of a fuzzy inference system is also influenced by the number of rules. Too few rules can restrict the model's expressive capability, while too many can introduce redundancy, overfitting, and reduced interpretability. To balance these factors, we incorporate a dynamic Rule Growing and Pruning (RG&RP) strategy into the ADAR framework. The RG&RP dynamically adjusts the rule base to maintain a balance between model complexity and representational capacity. When the existing rules are insufficient to cover complex regions of the input space or address persistent errors, the model introduces new rules to better represent these challenging data subsets. Conversely, rules that consistently show low importance—reflects by their rule importance weights—or become redundant after attribute pruning are removed to streamline the model's structure. The Rule Growing and Pruning (RG&RP) mechanism builds upon the learned rule importance weights $\{\beta_l\}$ from the Rule Weighting Mechanism. Instead of redefining these weights, RG&RP utilizes them alongside other performance metrics to dynamically adjust the rule base, ensuring efficient and effective adaptation.

**Rule Pruning Criterion:** A rule $l$ whose importance weight $\beta_l$ stays below a predefined threshold $\theta_r$ for an extended period is considered underutilized and may be removed. This pruning step helps the model's complexity with its actual requirements, retaining only those rules that contribute significantly to predictions.

Formally, if:

$$\beta_l < \theta_r \tag{5}$$

then rule $l$ is marked for removal, provided this condition holds over multiple training epochs.

**Rule Growing Criterion:** On the other hand, if the validation error shows insufficient improvement over a specific patience period—despite previous attribute pruning and training—this suggests that the existing rules may not sufficiently cover certain areas of data. In this case, the model adds a new rule to better cover the input space and improve predictive accuracy.

If:

$$\text{No improvement in validation error for } p \text{ epochs} \quad \text{and} \quad L < L_{\max}$$

(where $p$ is the patience parameter, and $L_{\max}$ is the maximum number of allowable rules), the model selects high-error samples from the training set to initialize a new rule's membership function parameters. The new rule aims to capture previously unexplained patterns and reduce residual errors.

### 3.7 Fuzzy Inference Module

In the ADAR framework, the Fuzzy Inference Module acts as the central computational layer, converting input data into relevant rule activations and generating the final prediction. While the weight-based mechanisms and dynamic structural adjustments focus on identifying the most important attributes and rules, the fuzzy inference process governs how these elements interact, enabling the model to effectively manage uncertainty, nonlinearity, and high-dimensional complexity.

The primary function of the Fuzzy Inference Module is to represent input features using fuzzy sets and combine them through fuzzy rules to generate reliable predictions. By employing parameterized membership functions (e.g., Gaussian functions) for each attribute in every rule, this module naturally effectively handles the gradual transitions and uncertainties inherent in real-world data. After calculating the rule firing strengths, they are aggregated and normalized to produce a crisp output. This approach ensures that the model integrates both the learned structure (rules and attributes) and the importance-guided weighting of these components, resulting in accurate and interpretable predictions.

Consider a set of $L$ fuzzy rules, each associated with a parameterized membership function for each input attribute. For rule $l$ and attribute $i$, let $\mu_{l,i}(x_i)$ denote the membership degree of the input $x_i$ to the fuzzy set defined by rule $l$. A common choice in ADAR is the Gaussian membership function:

$$\mu_{l,i}(x_i) = \exp\left(-\frac{(x_i - v_{l,i})^2}{2s_{l,i}^2}\right),$$ (6)

where $v_{l,i}$ and $s_{l,i}$ are the learnable center and width parameters of the Gaussian set.

For each rule $l$, the combined firing strength is computed by aggregating the membership degrees across all active attributes. Let $D$ be the number of attributes in a rule:

$$\text{Firing Strength}(R_l) = \prod_{i=1}^{D} \mu_{l,i}(x_i) \cdot \alpha_{l,i},$$ (7)

where $\alpha_{l,i}$ is the attribute importance weight, ensuring that less important attributes have a reduced influence on the rule's activation.

After computing firing strengths for all rules, they are weighted by the rule importance weights $\beta_l$, reflecting the overall importance of each rule:

$$\widetilde{f_l} = \text{Firing Strength}(R_l) \cdot \beta_l.$$ (8)

The normalized rule activation is then:

$$w_l = \frac{\widetilde{f_l}}{\sum_{m=1}^{L} \widetilde{f_m} + \epsilon},$$ (9)

where $\epsilon$ is a small constant to prevent division by zero.

Finally, each rule outputs a linear combination of selected attributes (after attribute pruning) as its consequent:

$$y_l = \sum_{i \in \mathcal{A}_l} c_{l,i} x_i,$$ (10)

where $\mathcal{A}_l$ is the set of active attributes in rule $l$, and $c_{l,i}$ are the consequent parameters. The final prediction is the weighted sum of the rule outputs:

$$y = \sum_{l=1}^{L} w_l y_l.$$ (11)

### 3.8 Integration with SOFENN

The integration of the ADAR framework with SOFENN follows similar adaptive training principles as those described in the ANFIS-based implementation, sharing core components including the Attribute Weighting Mechanism, Attribute Pruning (AP), Rule Weighting Mechanism, Rule Growing and Pruning (RG&RP), and the Fuzzy Inference Module. However, due to SOFENN's distinct structural characteristics, the integration exhibits notable differences in initialization strategies, rule evolution, and adaptive optimization approaches.

Specifically, the ADAR-SOFENN implementation initializes fuzzy rules similarly via clustering methods but emphasizes a more flexible, online-style structural adjustment. Unlike ANFIS, which predominantly refines a fixed set of fuzzy rules, SOFENN inherently supports incremental rule construction and pruning in a more dynamic, data-driven fashion. This aligns naturally with ADAR's dynamic attribute and rule adjustment capabilities, enhancing SOFENN's strengths in adaptively capturing evolving data distributions.

During training, the attribute-level importance weighting mechanism in SOFENN similarly guides periodic attribute pruning based on dynamically computed importance weights. Yet, distinctively, ADAR-SOFENN exploits SOFENN's inherently modular architecture to facilitate more frequent and granular attribute-level adjustments without incurring significant computational overhead or structural instability. Rule-level adaptations through RG&RP also differ slightly from ANFIS integration, as SOFENN allows incremental rule addition to occur continuously rather than at discrete intervals, thus enabling smoother adaptation to emerging patterns in data.

In both frameworks, attribute and rule pruning procedures include rigorous performance checks, reverting structural changes when excessive performance degradation is detected. However, ADAR-SOFENN uniquely benefits from SOFENN's incremental optimization strategy, allowing for more seamless integration of newly added or removed rules, typically requiring fewer complete reinitializations of the optimizer compared to ADAR-ANFIS.

Overall, ADAR-SOFENN capitalizes on SOFENN's dynamic and incremental structure, achieving an enhanced balance between model complexity and predictive performance, especially suited to datasets exhibiting non-stationary or evolving characteristics.

## 4 Experimental Design

The effectiveness of ADAR will be evaluated through a series of computational experiments. ADAR can be used with ANFIS [12] and SOFENN [16], a statistical online fuzzy extreme neural network. So their performances will be compared with the conventional ANFIS and SOFENN. Furthermore, we will also compare with the following models:

- **FuBiNFS [28]**: A novel Fuzzy Bilateral Neuro-Fuzzy System that enhances generalization and interpretability by constructing fuzzy rules in subspaces.
- **ANFIS-PSO [25]**: An ANFIS model integrated with Particle Swarm Optimization (PSO) to optimize parameters and improve prediction accuracy.
- **RVFL (Neuro-Fuzzy RVFL) [22]**: A model that integrates Random Vector Functional Link networks with Neuro-fuzzy systems, enhancing model transparency and generalization.
- **APLR [29]**: The Automatic Piecewise Linear Regression model is highly interpretable and serves as a key interpretability algorithm beyond the fuzzy logic domain.
- **DecisionTreeRegressor [3]**: A decision tree regression model is highly interpretable and serves as a representative algorithm beyond the fuzzy logic domain.
- **ADAR-ANFIS** and **ADAR-SOFENN**: The ANFIS and SOFENN models were enhanced using the ADAR framework proposed in this paper.

*4.0.1 Datasets.* We selected four datasets with different feature counts and complexities to cover a wide range of application scenarios:

- **Auto MPG [20]** (7 features): Used for predicting the miles per gallon (MPG) of automobiles.
- **Beijing PM2.5 [6]** (10 features): Used for predicting the concentration of PM2.5 in the air.
- **Boston Housing [10]** (13 features): Used for predicting the median house prices.
- **Appliances Energy Consumption [4]** (27 features): Used for predicting the energy consumption of household appliances.

Each dataset was subjected to basic data preprocessing steps, including handling missing values (if applicable) and standardization or normalization. These measures ensure consistent input scales and facilitate more reliable model comparisons.

*Note:* The structures of **APLR** and **DecisionTreeRegressor** do not involve rule quantity settings, so their performance remains unaffected by variations in the number of rules.

*4.0.2 Evaluation Metrics.* We utilized the following three metrics to assess both predictive performance and structural optimization of the models:

- **Root Mean Square Error (RMSE)**: RMSE measures the average deviation between the model's predicted values and the actual values. A lower RMSE indicates higher prediction accuracy. It is computed as:

$$\text{RMSE} = \sqrt{\frac{1}{N} \sum_{i=1}^{N} (\hat{y}_i - y_i)^2} \tag{12}$$

  where $N$ is the number of samples, $\hat{y}_i$ is the predicted value, and $y_i$ is the actual value.

- **Average Overlap Index ($I_{ov}$) [28]**: Measures the extent of overlap among rules in the fuzzy rule base. $I_{ov}$ computes the maximum overlapping area of Gaussian membership functions for all rules pairs under each attribute and averages the results across all attributes:

$$I_{ov}(L) = \frac{1}{D} \sum_{d=1}^{D} \left[ \max_{\substack{i,j=1 \\ i \neq j}}^{L} \frac{\int_{-\infty}^{+\infty} \min \left( \mu_{A_d}^{(l_i)}(x), \mu_{A_d}^{(l_j)}(x) \right) dx}{\min \left( \int_{-\infty}^{+\infty} \mu_{A_d}^{(l_i)}(x) dx, \int_{-\infty}^{+\infty} \mu_{A_d}^{(l_j)}(x) dx \right)} \right] \tag{13}$$

  where $D$ is the number of attributes, $L$ is the number of rules, and $\mu_{A_d}^{(l)}(x)$ denotes the membership function of the $l$-th rule on attribute $d$.

  A lower $I_{ov}$ value signifies reduced overlap between rules and greater distinguishability within the rule base, which enhances improve the model's generalization ability and interpretability by minimizing ambiguity among fuzzy rules.

- **Average Fuzzy Set Position Index ($I_{fsp}$) [14]**: $I_{fsp}$ evaluates the coverage of fuzzy sets by calculating the positional and shape differences between adjacent fuzzy sets :

$$I_{fsp}(L) = \frac{1}{LD} \sum_{d=1}^{D} \sum_{l=1}^{L-1} \left[ 2 \cdot \left( 0.5 - \phi_{A_d^{(l)}, A_d^{(l+1)}} + \psi_{A_d^{(l)}, A_d^{(l+1)}} \right) \right] \tag{14}$$

where

$$\phi_{A_d^{(l)}, A_d^{(l+1)}} = \exp \left[ -\frac{1}{2} \left( \frac{v_d^{(l)} - v_d^{(l+1)}}{s_d^{(l)} + s_d^{(l+1)}} \right)^2 \right] \tag{15}$$

$$\psi_{A_d^{(l)}, A_d^{(l+1)}} = \exp \left[ -\frac{1}{2} \left( \frac{v_d^{(l)} - v_d^{(l+1)}}{s_d^{(l)} - s_d^{(l+1)}} \right)^2 \right] \tag{16}$$

In these formulas, $A_d^{(l)}$ is the Gaussian fuzzy set of the $l$-th rule on attribute $d$, $v_d^{(l)}$ and $s_d^{(l)}$ represent the mean (center) and standard deviation (width) of the fuzzy set, respectively. The fuzzy sets are sorted by their means $v$, and $v_d^{(l)}$ and $v_d^{(l+1)}$ denote two consecutive neighboring fuzzy sets after sorting.

A lower $I_{fsp}$ value reflects more accurate positioning of fuzzy sets in the input space and more reasonable coverage, enhancing the model's interpretability and generalization ability.

*4.0.3 Model Configuration and Hyperparameters.* In all experiments, unless stated otherwise, the following hyperparameters were consistently set to maintain the comparability of results:

- **Learning Rate**: 0.01
- **Batch Size**: 512
- **Number of Training Iterations**: 1500

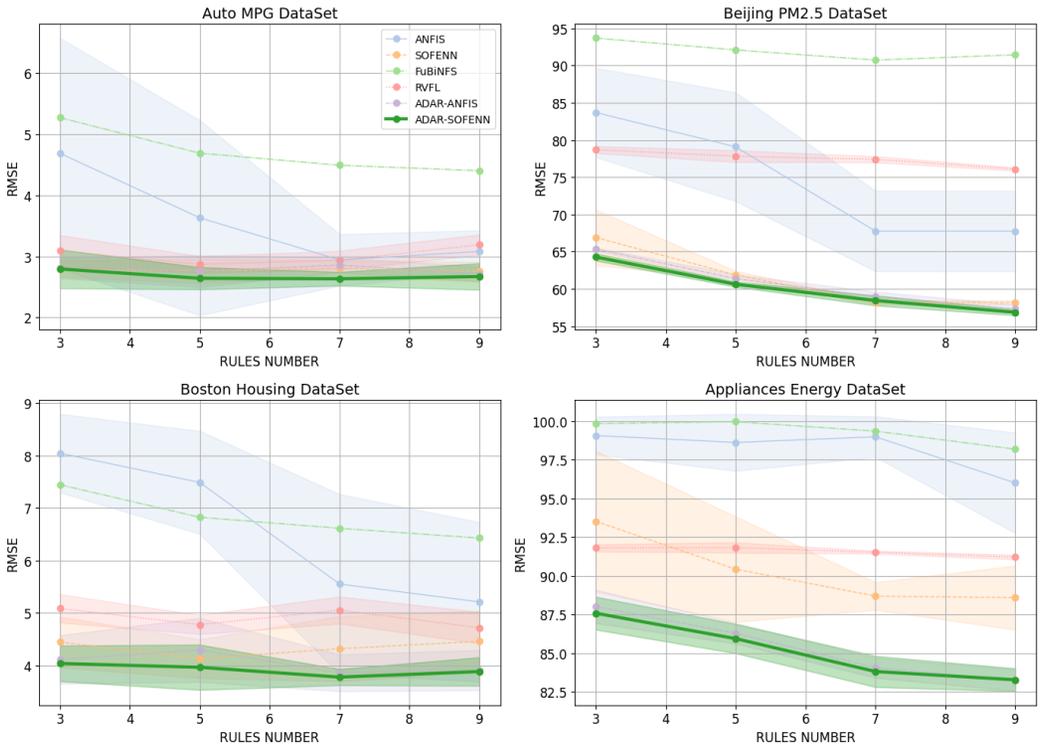

Fig. 2. Comparison of Interpretability Algorithms Across Datasets and Rule Numbers

For rule-based models, the maximum number of rules was tested at 3, 5, 7, and 9 to evaluate the impact of rule quantity on model performance.

## 4.1 Experimental Results and Discussion

In this section, we present and analyze the results from our various experiments. We begin with a broad comparison of different algorithms across multiple datasets to highlight the advantages of the ADAR framework. Next, We discuss the dynamic rule and attribute management mechanism of the ADAR framework, followed by the results of ablation study and, finally, a comprehensive parameter sensitivity analysis.

*Comparison of Algorithms Across Datasets.* Table 1 and Table 2 summarize the RMSE (Root Mean Square Error) performance of each algorithm across the four datasets. For rule-based models, the results show performance under different maximum rule quantities. Since the rule design of ANFIS-PSO differs from other fuzzy logic algorithms and the results for APLR and DecisionTreeRegressor are independent of rule quantity, their results are combined in Table 2.

*Advantages of the ADAR Framework on Complex Datasets.* The experimental results show that models built on the ADAR framework (ADAR-ANFIS and ADAR-SOFENN) consistently deliver better performance across all datasets, especially on complex datasets with a larger number of features. This highlights the ADAR framework's strong capability in handling high-dimensional data and intricate nonlinear relationships.

Table 1. RMSE Comparison Across Datasets with Varying Rules

| Dataset | Algorithm | RULE=3 | RULE=5 | RULE=7 | RULE=9 |
|---|---|---|---|---|---|
| Auto MPG (Variable=7) | ANFIS | 4.6879 ± 1.8879 | 3.6324 ± 1.5975 | 2.9424 ± 0.4225 | 3.0815 ± 0.3451 |
| | SOFENN | 2.8008 ± 0.1373 | 2.7614 ± 0.2003 | 2.8046 ± 0.1312 | 2.7513 ± 0.1600 |
| | FuBiNFS | 5.2729 ± 0.0000 | 4.6923 ± 0.0000 | 4.4941 ± 0.0000 | 4.4049 ± 0.0000 |
| | RVFL | 3.0897 ± 0.2582 | 2.8711 ± 0.1346 | 2.9358 ± 0.1559 | 3.1911 ± 0.1668 |
| | ADAR-ANFIS | 2.8108 ± 0.1675 | 2.7587 ± 0.2589 | 2.8580 ± 0.1219 | 2.7128 ± 0.1230 |
| | **ADAR-SOFENN** | **2.7948 ± 0.3171** | **2.6418 ± 0.1823** | **2.6332 ± 0.1122** | **2.6699 ± 0.2173** |
| Beijing PM2.5 (Variable=10) | ANFIS | 83.6791 ± 5.9950 | 79.0816 ± 7.3014 | 67.7683 ± 5.4045 | 67.7683 ± 5.4045 |
| | SOFENN | 66.9002 ± 3.7310 | 61.8837 ± 0.5297 | 58.1916 ± 0.2862 | 58.1916 ± 0.2862 |
| | FuBiNFS (2021) | 93.6840 ± 0.0000 | 92.1008 ± 0.0000 | 90.7340 ± 0.0000 | 91.4465 ± 0.0000 |
| | RVFL (2024) | 78.6992 ± 0.4646 | 77.8190 ± 0.8003 | 77.3965 ± 0.4112 | 76.0913 ± 0.2162 |
| | ADAR-ANFIS | 65.3408 ± 0.2019 | 61.3989 ± 0.5540 | 58.9481 ± 0.6897 | 57.3199 ± 0.5940 |
| | **ADAR-SOFENN** | **64.2827 ± 0.4883** | **60.6262 ± 0.3478** | **58.4496 ± 0.6825** | **56.8668 ± 0.4113** |
| Boston Housing (Variable=13) | ANFIS | 8.0455 ± 0.7479 | 7.4894 ± 0.9818 | 5.5524 ± 1.7124 | 5.2112 ± 1.5201 |
| | SOFENN | 4.4443 ± 0.4845 | 4.1290 ± 0.3700 | 4.3192 ± 0.6223 | 4.4596 ± 0.5837 |
| | FuBiNFS | 7.4459 ± 0.0000 | 6.8265 ± 0.0000 | 6.6156 ± 0.0000 | 6.4289 ± 0.0000 |
| | RVFL | 5.0860 ± 0.2700 | 4.7784 ± 0.1817 | 5.0550 ± 0.2580 | 4.7192 ± 0.3001 |
| | ADAR-ANFIS | 4.1125 ± 0.4632 | 4.2964 ± 0.6100 | 3.8586 ± 0.3517 | 3.9099 ± 0.3828 |
| | **ADAR-SOFENN** | **4.0378 ± 0.3398** | **3.9641 ± 0.4319** | **3.7770 ± 0.1541** | **3.8831 ± 0.2713** |
| Appliances Energy (Variable=27) | ANFIS | 99.0577 ± 1.2197 | 98.6092 ± 1.8407 | 98.9762 ± 1.3135 | 95.9975 ± 3.2503 |
| | SOFENN | 93.5092 ± 4.5456 | 90.4177 ± 3.3490 | 88.6798 ± 0.8949 | 88.5745 ± 2.0701 |
| | FuBiNFS | 99.8365 ± 0.0000 | 99.9557 ± 0.0000 | 99.3492 ± 0.0000 | 98.1800 ± 0.0000 |
| | RVFL | 91.7919 ± 0.2311 | 91.8162 ± 0.3387 | 91.5115 ± 0.0933 | 91.1958 ± 0.1544 |
| | ADAR-ANFIS | 88.0047 ± 1.0725 | 86.2727 ± 0.6647 | 84.0432 ± 0.6443 | 83.2472 ± 0.6939 |
| | **ADAR-SOFENN** | **87.5710 ± 1.0646** | **85.9295 ± 0.9476** | **83.8022 ± 1.0049** | **83.2657 ± 0.7424** |

Table 2. Best RMSE Comparison Across Algorithms and Datasets

| Dataset | Algorithm | Best RMSE | Rule |
|---|---|---|---|
| Auto MPG (Variable=7) | ANFIS-PSO | 2.8578 ± 0.1956 | 14 |
| | APLR | 2.7572 ± 0.1336 | - |
| | DecisionTreeRegressor | 3.6448 ± 0.4722 | - |
| | ADAR-ANFIS | 2.7128 ± 0.1230 | 9 |
| | **ADAR-SOFENN** | **2.6332 ± 0.1122** | **7** |
| Beijing PM2.5 (Variable=10) | ANFIS-PSO | 85.0095 ± 1.3883 | 20 |
| | APLR | 57.11896 ± 0.1718 | - |
| | DecisionTreeRegressor | 55.81432 ± 0.9817 | - |
| | ADAR-ANFIS | 57.3199 ± 0.5940 | 9 |
| | **ADAR-SOFENN** | **56.8668 ± 0.4113** | **9** |
| Boston Housing (Variable=13) | ANFIS-PSO | 8.6875 ± 0.2733 | 26 |
| | **APLR** | **3.4227 ± 0.4218** | **-** |
| | DecisionTreeRegressor | 5.1525 ± 0.4114 | - |
| | ADAR-ANFIS | 3.8586 ± 0.3517 | 7 |
| | ADAR-SOFENN | 3.7770 ± 0.1541 | 7 |
| Appliances Energy (Variable=27) | ANFIS-PSO | 100.2024 ± 0.2374 | 54 |
| | APLR | 83.9781 ± 1.1202 | - |
| | DecisionTreeRegressor | 98.7506 ± 2.3485 | - |
| | **ADAR-ANFIS** | **83.2472 ± 0.6939** | **9** |
| | ADAR-SOFENN | 83.2657 ± 0.7424 | 9 |

*Comparison with Traditional Models.* In comparison to conventional ANFIS and SOFENN models, the ADAR framework improves prediction accuracy by incorporating attribute and rule weighting mechanism that automatically detect and prioritize important features and rules. For example, in the Appliances Energy Consumption dataset, ADAR-SOFENN achieved an RMSE of 83.2657 with 9 rules (RULE=9), significantly outperforming traditional models. This enhancement in performance is largely due to the ADAR framework's ability to dynamically adjust attributes and rules, allowing the model to better capture critical patterns in the data.

*Comparison with Advanced Fuzzy Logic Models.* When compared to advanced fuzzy logic models like FuBiNFS and ANFIS-PSO, these models, while showing better results in some cases, generally fall short of the performance seen with ADAR framework models. FuBiNFS improves generalization by creating fuzzy rules within subspaces but lacks dynamic weight assignment for feature and rule significance, limiting its performance on high-dimensional datasets. ANFIS-PSO uses the Particle Swarm Optimization algorithm to globally optimize model parameters, leading to some performance improvements. However, its fixed structure restricts the model's ability to adapt dynamically to the complexity of the data.

*Comparison with Interpretability Algorithms.* Interpretability algorithms beyond the fuzzy logic domain, such as APLR and DecisionTreeRegressor, showed strong performance on specific datasets. For instance, APLR achieved an RMSE of 3.4227 on the Boston Housing dataset, and Decision-TreeRegressor attained an RMSE of 55.8143 on the Beijing PM2.5 dataset. However, in most cases, models based on the ADAR framework either outperformed or matched these algorithms while offering superior interpretability.

## 4.2 ADAR Framework's Rule and Attribute Management

We evaluate the ADAR framework in combination with ANFIS on the Appliances Energy Consumption dataset to analyze the dynamic processes of rule growth, rule pruning, and attribute pruning during training. The model begins with two fuzzy rules and undergoes training for 500 epochs with a batch size of 64. The thresholds for attribute pruning and rule pruning are set at 0.1 and 0.25, respectively. This approach aims to simplify the model while preserving or improving its predictive performance.

*Results and Analysis.* In the early training phase (around epochs 50 to 150), the model detects previously unlearned data patterns and dynamically introduces new rules, increasing the rule count from the initial 2 to 14. This growth coincides with a notable reduction in validation loss, dropping from approximately 1.1 to 0.7. The newly added rules effectively capture high-error data points and refine decision boundaries, enhancing the model's predictive accuracy.

At the same time, the model utilizes an attribute weighting mechanism to dynamically prune irrelevant input features. Attributes with importance weights below 0.1 are gradually deactivated, minizing redundancy in each rule's input. By epoch 200, most irrelevant attributes have been removed, leaving an average of approximately 15 active attributes per rule. Notably, this pruning does not negatively impact validation loss; instead, it enhances the model's generalization ability.

In the later training phase (after epoch 350), a rule pruning mechanism is applied to enhance interpretability and computational efficiency. Rules with importance weights below 0.25 are eliminated, reducing the active rule count from 14 to 10. Despite this reduction, the validation loss remains stable, indicating that the model effectively removes redundant rules while retaining its essential decision-making capabilities.

Figure 3 illustrates the dynamic variations in validation loss and rule count during training. The solid line represents the trend in validation loss, while the dashed line shows the total number of rules in the model. During the early training stages, validation loss drops sharply as new rules are introduced. As the model transitions into the fine-tuning phase, both the rule count and validation loss stabilize, indicating the completion of the dynamic adjustment process.

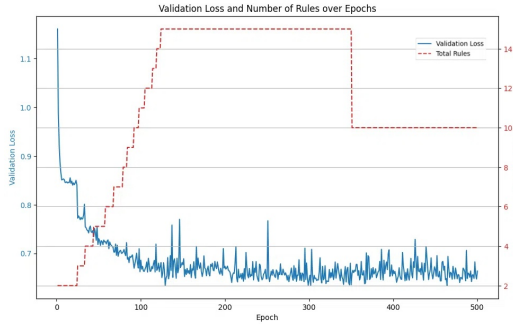

Fig. 3. Dynamic Rule Growth and Validation Loss Convergence in the ADAR Framework

## 4.3 Ablation Study Results Analysis

To comprehensively evaluate the impact of each component within the Adaptive Dynamic Attribute and Rule (ADAR) framework on model performance and structural optimization, we conducted a systematic ablation study on two fuzzy neural network models based on the ADAR framework: ADAR-SOFENN (Self-Organizing Fuzzy Neural Network with Adaptive Dynamic Attributes and Rules) and ADAR-ANFIS (Adaptive Dynamic Attribute and Rule-based ANFIS). This study aims to quantify the contributions of the Rule Growing and Pruning (RG&RP) and Attribute Pruning (AP) components to the models, with a particular emphasis on the role of weighting weighting mechanisms in attribute selection and rule refinement.

*Experimental Design and Configuration.* The experiments utilized the Beijing PM2.5 dataset from the UCI Machine Learning Repository. Ten feature variables were selected: Year, Month, Day, Hour, Dew Point Temperature (DEWP), Temperature (TEMP), Pressure (PRES), Cumulative Wind Speed (Iws), Cumulative Hours (Is), and Cumulative Precipitation (Ir), with PM2.5 concentration as the target variable. To maintain data quality, missing values were addressed, and the dataset was standardized.

The dataset was divided into an 80% training set and a 20% test set. The training set was further split into an 80% training subset and a 20% validation set for model training and evaluation. The model was trained with the following parameters: 1500 epochs, a batch size of 512, aa learning rate of 0.01, attribute pruning every 25 epochs, and pruning thresholds of 0.1 for attributes and 0.25 for rules.

For model configurations, we established four distinct experimental setups for each model:

- **Baseline Model (SOFENN and ANFIS)**: This setup excludes optimization components, with both RG & RP and AP disabled, and no weighting mechanisms incorporated. The model relies on fixed rules and attributes without adaptive adjustment capabilities.
- **Baseline Model + RG&RP**: Only the Rule Growing and Pruning component enabled, while the Attribute Pruning component disabled. This configuration incorporates a **rule weighting mechanism**, allowing the model to dynamically adjust the importance and number of rules. By learning rule importance weights, the model can identify and remove less significant rules while introducing new ones as needed to adapt to data complexity.

- **Baseline Model + AP**: Only the Attribute Pruning (AP) component enabled, while the Rule Growing and Pruning (RG&RP) component remains disabled. This configuration incorporates an **attribute weighting mechanism** to dynamically adjust the relevance and weights of input features. By learning attribute importance weights, the model can prioritize important attributes while suppressing redundant or less relevant features, enhancing generalization performance.
- **Full Model (Baseline Model + RG&RP + AP, i.e., ADAR-SOFENN and ADAR-ANFIS)**: Both the Rule Growing and Pruning (AP) and Attribute Pruning (RG&RP) components are enabled. This setup integrates rule weighting and attribute weighting mechanisms, allowing for comprehensive model structure optimization.

Each configuration was trained with maximum rule limits of 3, 5, 7, and 9, with each experiment repeated five times to ensure reliability and statistical significance. When the RG&RP component is enabled, the model can dynamically add or remove rules but will not exceed the predefined maximum. For evaluation, We use **RMSE**, $I_{ov}$, and $I_{fsp}$ as defined in Evaluation Metrics. A lower $I_{ov}$ indicates reduced rule overlap, enhancing interpretability. While a lower $I_{fsp}$ implies more precise fuzzy set positioning, improving coverage and interpretability.

*Experimental Results and Analysis.* Tables 3 and Table 4 present the ablation results for different max rule numbers.

Table 3. Ablation Study Results of ADAR-SOFENN

| Model Configuration | Max Rules | RMSE (±Std) | $I_{ov}$ (±Std) | $I_{fsp}$ (±Std) |
|---|---|---|---|---|
| **SOFENN** | 3 | 66.9002 ± 3.7310 | 2.3862 ± 0.3029 | 0.8611 ± 0.1099 |
| | 5 | 61.8837 ± 0.5297 | 2.7036 ± 0.3439 | 0.8968 ± 0.0962 |
| | 7 | 60.1431 ± 0.9294 | 3.1246 ± 0.3581 | 1.0148 ± 0.0377 |
| | 9 | 58.1916 ± 0.2862 | 3.8300 ± 0.5557 | 1.0067 ± 0.0707 |
| **SOFENN + RG&RP** | 3 | 65.7658 ± 2.3611 | 2.8810 ± 0.3328 | 0.7608 ± 0.0554 |
| | 5 | 61.7878 ± 0.5831 | 2.6595 ± 0.1973 | 0.9107 ± 0.0697 |
| | 7 | 59.0466 ± 0.5816 | 3.1668 ± 0.4685 | 0.9276 ± 0.0739 |
| | 9 | 58.0395 ± 0.7897 | 3.0303 ± 0.1364 | 0.9932 ± 0.0569 |
| **SOFENN + AP** | 3 | 64.7029 ± 0.5744 | 1.4909 ± 0.1728 | 0.4680 ± 0.1428 |
| | 5 | 61.1893 ± 0.3582 | 1.5838 ± 0.2418 | 0.6776 ± 0.1207 |
| | 7 | 59.4569 ± 0.4296 | 1.8413 ± 0.2813 | 0.6461 ± 0.1128 |
| | 9 | 57.6671 ± 0.3984 | 1.9144 ± 0.3762 | 0.7434 ± 0.1471 |
| **ADAR-SOFENN** | 3 | 64.2827 ± 0.4883 | 0.5263 ± 0.0756 | 0.4028 ± 0.0340 |
| | 5 | 60.6262 ± 0.3478 | 0.8624 ± 0.0246 | 0.6639 ± 0.0808 |
| | 7 | 58.4496 ± 0.6825 | 0.9425 ± 0.0129 | 0.6326 ± 0.0380 |
| | 9 | 57.2381 ± 0.4417 | 0.9676 ± 0.0076 | 0.7978 ± 0.0603 |

*Performance of Baseline Models.* The baseline models (SOFENN and ANFIS) exhibit relatively high RMSE values across all maximum rule numbers, with RMSE gradually decreasing as the maximum rule number increases. For instance, the RMSE of ADAR-SOFENN decreases from 66.90 ± 3.73 at a maximum rule number of 3 to 58.19 ± 0.29 at a maximum rule number of 9. Similarly, ADAR-ANFIS sees a reduction from 88.18 ± 5.39 to 66.93 ± 3.58. This indicates that increasing the number of rules can enhance predictive performance to some extent. However, the $I_{ov}$ values of the baseline models remain relatively high, especially for ANFIS, where $I_{ov}$ rises from 12.45 ± 10.26 at a maximum rule number of 3 to 24.32 ± 6.53 at a maximum rule number of 9. This suggests significant rule overlap and redundancy in the baseline models, which negatively affects generalization ability and interpretability.

Table 4. Ablation Study Results of ADAR-ANFIS

| Model Configuration | Max Rules | RMSE (±Std) | $I_{ov}$ (±Std) | $I_{fsp}$ (±Std) |
|---|---|---|---|---|
| **ANFIS** | 3 | 88.1843 ± 5.3894 | 12.4502 ± 10.2629 | 0.7066 ± 0.0818 |
| | 5 | 79.0816 ± 7.3014 | 20.0718 ± 3.9651 | 0.7989 ± 0.0364 |
| | 7 | 70.3829 ± 7.7170 | 27.8611 ± 7.4576 | 0.8690 ± 0.0404 |
| | 9 | 66.9279 ± 3.5835 | 24.3194 ± 6.5254 | 0.9146 ± 0.0238 |
| **ANFIS + RG&RP** | 3 | 66.8949 ± 2.1634 | 2.8815 ± 0.9386 | 0.8243 ± 0.1218 |
| | 5 | 62.7235 ± 1.1389 | 3.1181 ± 0.2536 | 0.9718 ± 0.0584 |
| | 7 | 60.6018 ± 0.7743 | 5.2207 ± 1.5654 | 1.0168 ± 0.0866 |
| | 9 | 59.5692 ± 0.8444 | 6.6771 ± 2.3137 | 1.0322 ± 0.0576 |
| **ANFIS + AP** | 3 | 65.9599 ± 0.6826 | 1.2917 ± 0.2949 | 0.4769 ± 0.1595 |
| | 5 | 62.7266 ± 0.7535 | 1.7280 ± 0.2759 | 0.6823 ± 0.1729 |
| | 7 | 60.7769 ± 0.6910 | 1.8385 ± 0.3599 | 0.8217 ± 0.0807 |
| | 9 | 58.6765 ± 0.7500 | 2.0903 ± 0.3281 | 0.8006 ± 0.0971 |
| **ADAR-ANFIS** | 3 | 65.3408 ± 0.2019 | 0.6854 ± 0.1114 | 0.4727 ± 0.0925 |
| | 5 | 61.3989 ± 0.5540 | 0.8742 ± 0.0754 | 0.5872 ± 0.1514 |
| | 7 | 58.9481 ± 0.6897 | 0.9447 ± 0.0181 | 0.7257 ± 0.1630 |
| | 9 | 57.3199 ± 0.5940 | 0.9503 ± 0.0144 | 0.7253 ± 0.1146 |

*Impact of Adding RG&RP Component.* In configurations where only the Rule Growing and Pruning (RG&RP) component is enabled, RMSE values decrease significantly, highlighting the positive impact of RG&RP component on predictive accuracy. For example, ADAR-SOFENN + RG&RP sees its RMSE drop from 66.90 ± 3.73 to 65.77 ± 2.36 at a maximum rule number of 3, showcasing the effectiveness of the RG&RP component in optimizing predictive performance. Similarly, ADAR-ANFIS + RG&RP shows an even more substantial improvement, with RMSE decreasing from 88.18 ± 5.39 to 66.89 ± 2.16 under the same conditions, further confirming the RG&RP component's ability to significantly reduce prediction errors.

In the ADAR-SOFENN + RG&RP configuration, at a maximum rule number of 3, the $I_{ov}$ value shows a slight increases from 2.3862 ± 0.3029 to 2.8810 ± 0.3328. This slight increase can be attributed to the already low $I_{ov}$ value in the SOFENN baseline model, which indicates minimal rule overlap. With RG&RP enabled, the model may have added new rules to better fit the data. The addition of these rules results in some overlap with existing ones, leading to the slight rise in $I_{ov}$ value. However, this increase is minimal, and the $I_{ov}$ value remains low, indicating that the rule overlap is still controllable.

In contrast, in the ADAR-ANFIS + RG&RP configuration, at a maximum rule number of 3, the $I_{ov}$ value decreases significantly from 12.4502 ± 10.2629 to 2.8815 ± 0.9386. This is due to the ANFIS baseline model exhibiting a high degree of rule overlap and low rule distinguishability. With the RG&RP component enabled, the model effectively reduces overlap by pruning redundant rules and adjusting rule parameters, leading to a sustainable decrease in the $I_{ov}$ value. This indicates that the RG&RP component has a more sustainable impact on optimizing the rule base structure in the ANFIS model.

Therefore, the impact of the RG&RP component differs across models: for SOFENN, which has an initial low $I_{ov}$ value, the $I_{ov}$ value slightly increases due to minor overlap introduced by new rules; for ANFIS, with its initially high $I_{ov}$ value, the RG&RP component effectively reduces rule overlap, leading to a significant decrease in $I_{ov}$ value.

Additionally, the RG&RP component optimizes the $I_{fsp}$ metric across various maximum rule numbers. For instance, in ADAR-SOFENN + RG&RP, $I_{fsp}$ decreases from 0.8611 ± 0.1099 to 0.7608 ± 0.0554 at a maximum rule number of 3, indicating improved fuzzy set positioning. These enhancements demonstrate that the RG&RP component, by adjusting rule numbers and parameters, improves the distribution accuracy of fuzzy sets in the input space, ultimately enhancing the model's generalization ability and predictive performance.

*Impact of Adding AP Component.* In the configurations where only the Attribute Pruning (AP) component is enabled, both the RMSE and $I_{ov}$ values of the models decrease significantly, and the $I_{fsp}$ metric is effectively optimized. For example, ADAR-SOFENN + AP achieves an RMSE of 64.70 ± 0.57, an $I_{ov}$ of 1.4909 ± 0.1728, and a $I_{fsp}$ significantly reduced to 0.4680 ± 0.1428 at a maximum rule number of 3. This demonstrates that the attribute pruning mechanism, supported by the attribute weighting mechanism, successfully suppresses irrelevant features, improving the model's simplicity and generalization ability. In the case of ADAR-ANFIS + AP, RMSE decreases from 88.18 ± 5.39 to 65.96 ± 0.68, $I_{ov}$ from 12.45 ± 10.26 to 1.2917 ± 0.2949, and $I_{fsp}$ from 0.7066 ± 0.0818 to 0.4769 ± 0.1595 under the same conditions, highlighting the significant impact of the AP component on model performance.

The attribute weighting mechanism enables the model to dynamically adjust the importance of attributes by assigning weights to the input attributes in each rule. This helps suppress redundant or irrelevant features, improving the positioning accuracy of fuzzy sets in the input space, further optimizing the $I_{fsp}$ metric. As a result, the model's predictive accuracy is enhanced, while its interpretability and reliability are also improved.

*Advantages of the Full Model.* In the full model configurations (ADAR-SOFENN and ADAR-ANFIS), both the Rule Growing and Pruning (RG&RP) and Attribute Pruning (AP) components are enabled, resulting in optimal performance across all metrics. This demonstrates the synergistic effect of the RG&RP and AP components, which together provide a comprehensive optimization of the model structure and its performance. Specifically, the rule weighting mechanism dynamically refines the rule base structure, minimizing redundancy and excessive overlap, while the attribute weighting mechanism selectively adjusts input features, simplifying the model structure and improving the positioning accuracy of fuzzy sets.

The full models consistently demonstrate low RMSE, low $I_{ov}$, and low $I_{fsp}$ values across all maximum rule numbers, indicating significant improvements in both predictive accuracy and the optimization of rule base structure and fuzzy set positioning. For example, ADAR-SOFENN achieves an RMSE of 57.24 ± 0.44, $I_{ov}$ of 0.9676 ± 0.0076, and $I_{fsp}$ of 0.7978 ± 0.0603 at a maximum rule number of 9. Similarly, ADAR-ANFIS achieves an RMSE of 57.32 ± 0.59, $I_{ov}$ of 0.9503 ± 0.0144, and $I_{fsp}$ of 0.7253 ± 0.1146 under the same conditions. These results clearly demonstrate the advantages of the ADAR framework in optimizing rule base, maintaining low rule overlap, and ensuring optimized fuzzy set positioning across different rule numbers, significantly enhancing the model's predictive accuracy and generalization ability.

## 4.4 Parameter Sensitivity Analysis

To explore the impact of hyperparameters on the ADAR framework, we conducted a parameter sensitivity analysis on ADAR-ANFIS by varying thresholds and pruning frequencies. The goal of this analysis was to assess how different parameter configurations influence the model's predictive accuracy and interpretability, ultimately identifying the optimal parameter combination for improved model performance. Table 5 summarizes the average results for each parameter setting across three independent experiments.

*Experimental Setup for Sensitivity.* The sensitivity analysis examines five critical hyperparameters of the ADAR-ANFIS model:

- **Growth Threshold (G_Thres)**: Regulates the sensitivity of the rule-growing mechanism, with values tested at $5 \times 10^{-5}$ and $1 \times 10^{-4}$.
- **Prune Rule Threshold (PR_Thres)**: Determines the threshold for pruning rules based on importance weights, with values of 0.05 and 0.1.
- **Prune Rule Frequency (PR_Freq)**: Defines the interval (in epochs) between rule pruning operations, with testing conducted at 50 and 100 epochs.
- **Prune Attribute Threshold (PA_Thres)**: Establishes the threshold for pruning attributes within rules, with tested values of 0.05 and 0.1.
- **Prune Attribute Frequency (PA_Freq)**: Specifies the frequency (in epochs) for attribute pruning, with 25 and 50 epochs considered.

A total of 32 unique configurations were tested, each representing a combination of the aforementioned parameters. For every configuration, the model was trained and evaluated independently three times to account for random variations, with the results averaged.

*Results and Analysis.* The influence of hyperparameters on model performance was assessed using the following metrics:

- **Average Test RMSE**: The root mean squared error on the test set, indicating the model's predictive accuracy. Lower values are preferred.
- **Average $I_{ov}$**: The mean overlap index of fuzzy sets, reflecting the extent of overlap between fuzzy membership functions. **Lower $I_{ov}$ values are better**, indicating reduced overlap between rules and better discriminability, which enhances the model's generalization ability and interpretability.
- **Average $I_{fsp}$**: The mean fuzzy set position index, representing the dispersion of fuzzy set centers. **Lower $I_{fsp}$ values are better**, indicating more accurate positioning of fuzzy sets in the input space, leading to better coverage and improved interpretability and generalization.
- **Average Final Rules**: The average number of rules in the final model.
- **Average Final Attributes**: The average total number of attributes included in all rules.

*Impact of Growth Threshold.* When comparing the two growth thresholds $5 \times 10^{-5}$ and $1 \times 10^{-4}$, the results show that a lower growth threshold generally leads to models with slightly lower Average Test RMSE and final rules. For example, the parameter combination `G5e-05_PR0.1_PF50_PA0.05_PAF25` achieved the lowest Average Test RMSE of 0.8034, with an average of 9.00 final rules. This indicates that a lower growth threshold enables the model to generate more rules, allowing it to capture more complex patterns in the data, thus improving predictive performance.

However, a lower growth threshold may also lead to higher $I_{ov}$ and $I_{fsp}$ values, indicating increased overlap between rules and less accurate positioning of fuzzy sets. For instance, configurations in the `G5e-05` series have $I_{ov}$ values ranging from 0.64 to 0.68, and $I_{fsp}$ values between 0.58 and 0.67. Higher $I_{ov}$ and $I_{fsp}$ values may reduce the model's discriminability and interpretability.

Table 5. Parameter Sensitivity Experiment Results - Mean Values of Three Repeats

| Param_Set | Average_Test_RMSE | Average_$I_{ov}$ | Average_$I_{fsp}$ | Average_Final_Rules | Average_Final_Attributes |
|---|---|---|---|---|---|
| G5e−05_PR0.05_PF50_PA0.05_PAF25 | 0.816705 | 0.659805 | 0.581153 | 7.0 | 189.0 |
| G5e−05_PR0.05_PF50_PA0.05_PAF50 | 0.821961 | 0.674249 | 0.536936 | 7.0 | 189.0 |
| G5e−05_PR0.05_PF50_PA0.1_PAF25 | 0.810586 | 0.679352 | 0.577585 | 9.33 | 249.0 |
| G5e−05_PR0.05_PF50_PA0.1_PAF50 | 0.809774 | 0.640174 | 0.636667 | 8.67 | 232.0 |
| G5e−05_PR0.05_PF100_PA0.05_PAF25 | 0.822867 | 0.668093 | 0.577141 | 6.67 | 179.67 |
| G5e−05_PR0.05_PF100_PA0.05_PAF50 | 0.833914 | 0.661899 | 0.584540 | 6.67 | 179.67 |
| G5e−05_PR0.05_PF100_PA0.1_PAF25 | 0.816463 | 0.629675 | 0.633326 | 8.33 | 222.33 |
| G5e−05_PR0.05_PF100_PA0.1_PAF50 | 0.825841 | 0.603897 | 0.667261 | 5.33 | 141.33 |
| G5e−05_PR0.1_PF50_PA0.05_PAF25 | 0.803431 | 0.644519 | 0.630471 | 9.0 | 241.33 |
| G5e−05_PR0.1_PF50_PA0.05_PAF50 | 0.825825 | 0.627980 | 0.642903 | 6.0 | 162.0 |
| G5e−05_PR0.1_PF50_PA0.1_PAF25 | 0.819882 | 0.655016 | 0.591918 | 7.33 | 192.67 |
| G5e−05_PR0.1_PF50_PA0.1_PAF50 | 0.802829 | 0.673278 | 0.604944 | 10.33 | 278.0 |
| G5e−05_PR0.1_PF100_PA0.05_PAF25 | 0.811259 | 0.600041 | 0.719270 | 6.67 | 178.0 |
| G5e−05_PR0.1_PF100_PA0.05_PAF50 | 0.826052 | 0.659374 | 0.583199 | 6.33 | 170.33 |
| G5e−05_PR0.1_PF100_PA0.1_PAF25 | 0.805895 | 0.647840 | 0.656270 | 9.33 | 248.67 |
| G5e−05_PR0.1_PF100_PA0.1_PAF50 | 0.823679 | 0.606701 | 0.674123 | 6.33 | 167.33 |
| G0.0001_PR0.05_PF50_PA0.05_PAF25 | 0.825171 | 0.612779 | 0.668276 | 7.33 | 183.33 |
| G0.0001_PR0.05_PF50_PA0.05_PAF50 | 0.831590 | 0.669550 | 0.553278 | 4.67 | 125.67 |
| G0.0001_PR0.05_PF50_PA0.1_PAF25 | 0.816230 | 0.639937 | 0.661261 | 7.67 | 204.0 |
| G0.0001_PR0.05_PF50_PA0.1_PAF50 | 0.800953 | 0.669415 | 0.604870 | 10.33 | 278.0 |
| G0.0001_PR0.05_PF100_PA0.05_PAF25 | 0.833910 | 0.660353 | 0.558622 | 6.0 | 162.0 |
| G0.0001_PR0.05_PF100_PA0.05_PAF50 | 0.815831 | 0.629507 | 0.670356 | 7.0 | 188.67 |
| G0.0001_PR0.05_PF100_PA0.1_PAF25 | 0.846509 | 0.599343 | 0.616923 | 3.67 | 95.67 |
| G0.0001_PR0.05_PF100_PA0.1_PAF50 | 0.806566 | 0.619445 | 0.689826 | 8.0 | 211.33 |
| G0.0001_PR0.1_PF50_PA0.05_PAF25 | 0.812810 | 0.647198 | 0.627898 | 11.33 | 305.33 |
| G0.0001_PR0.1_PF50_PA0.05_PAF50 | 0.824287 | 0.647859 | 0.628169 | 7.67 | 206.33 |
| G0.0001_PR0.1_PF50_PA0.1_PAF25 | 0.816196 | 0.632600 | 0.662316 | 10.0 | 264.0 |
| G0.0001_PR0.1_PF50_PA0.1_PAF50 | 0.829588 | 0.631592 | 0.594863 | 6.67 | 178.67 |
| G0.0001_PR0.1_PF100_PA0.05_PAF25 | 0.812394 | 0.617858 | 0.681755 | 6.33 | 170.67 |
| G0.0001_PR0.1_PF100_PA0.05_PAF50 | 0.812964 | 0.628225 | 0.655215 | 8.0 | 215.0 |
| G0.0001_PR0.1_PF100_PA0.1_PAF25 | 0.819645 | 0.649383 | 0.616592 | 7.67 | 205.33 |
| G0.0001_PR0.1_PF100_PA0.1_PAF50 | 0.826528 | 0.594279 | 0.681473 | 5.33 | 143.0 |

On the other hand, a higher growth threshold ($10^{-4}$) tends to produce fewer rules, with relatively lower $I_{ov}$ and $I_{fsp}$ values. For example, the configuration G0.0001_PR0.05_PF100_PA0.1_PAF25 has an $I_{ov}$ of 0.5993, $I_{fsp}$ of 0.6169, and only 3.67 rules. However, a too high growth threshold may limit the model's ability to learn complex relationships, leading to a decrease in predictive accuracy, with an Average Test RMSE of 0.8465. This suggests that selecting the appropriate growth threshold requires balancing **predictive performance** with $I_{ov}$ and $I_{fsp}$ values.

*Impact of Prune Rule Threshold and Frequency.* Higher prune rule thresholds (0.1) typically results in a reduced number of rules in the model, as more rules are pruned. However, this does not always lead to better performance. For example, the Average Test RMSE of the parameter combination G5e−05_PR0.05_PF50_PA0.05_PAF25 is 0.8167, which is slightly higher than 0.8034 of G5e−05_PR0.1_PF50_PA0.05_PAF25. This could be because a higher prune threshold reduces model complexity while retaining more important rules, which in turn lower $I_{ov}$ and $I_{fsp}$ values. Specifically, configurations with PR_Thres = 0.1 generally have lower $I_{ov}$ and $I_{fsp}$ values, improving the model's discriminability and interpretability.

Regarding prune rule frequency, **more frequent pruning** (PR_Freq = 50) helps to quickly eliminate less important rules, leading to a further reduction in $I_{ov}$ and $I_{fsp}$ values, which enhances the model's generalization ability. In terms of RMSE performance, configurations with a pruning frequency of 50 epochs tend to show slightly lower RMSE values compared to those with a pruning frequency of 100 epochs.

*Impact of Prune Attribute Threshold and Frequency.* The prune attribute threshold significantly affects model complexity as well as the $I_{ov}$ and $I_{fsp}$ values. Higher prune attribute thresholds (0.1) reduce the number of attributes in the model while lowering $I_{ov}$ and $I_{fsp}$ values. For example, configurations with PA_Thres = 0.1 generally exhibits lower $I_{ov}$ and $I_{fsp}$ values, indicating reduced overlap between fuzzy sets and more accurate positioning. However, the effect on RMSE performance is mixed. In some instances, higher prune attribute thresholds can slightly increase RMSE, suggesting that excessive attribute pruning may remove important input features. Likewise, the influence of prune attribute frequency should be taken into account. More frequent attribute pruning (PAF = 25) helps quickly eliminate less important attributes, leading to reduced $I_{ov}$ and $I_{fsp}$ values. However, the impact on RMSE performance is inconsistent, highlighting the need for a balance between pruning frequency and predictive accuracy.

*Interaction Between Parameters.* Taking all parameters into account, the combination of a low growth threshold with a moderate prune rule threshold and an appropriate prune attribute threshold can maintain good predictive performance while reducing $I_{ov}$ and $I_{fsp}$ values. For example, the parameter combination `G5e-05_PR0.1_PF50_PA0.1_PAF25` achieved an Average Test RMSE of 0.8199, $I_{ov}$ of 0.6550, $I_{fsp}$ of 0.5919, and an average of 7.33 rules. This demonstrates that an optimal parameter mix can balance predictive performance and model interpretability. However, it is important to note that excessively reducing $I_{ov}$ and $I_{fsp}$ values (i.e., excessively reducing rule overlap and fuzzy set dispersion) could prevent the model from capturing complex data patterns, potentially impacting predictive performance. Therefore, finding the right balance between these parameters is essential for simultaneously optimizing RMSE, $I_{ov}$, and $I_{fsp}$ values.

*Overlap Index and Fuzzy Set Position Index.* While lower $I_{ov}$ and $I_{fsp}$ values enhance the model's discriminability and interpretability, it is essential to strike a balance between reducing $I_{ov}$ and $I_{fsp}$ values and maintaining good predictive performance in practical applications. The experimental results show that configurations with excessively low $I_{ov}$ and $I_{fsp}$ values (e.g., `G0.0001_PR0.1_PF100_PA0.1_PAF50`, where $I_{ov} = 0.5943$ and $I_{fsp} = 0.6815$) may have predictive performance (RMSE = 0.8265) that is that is inferior to those with more moderate $I_{ov}$ and $I_{fsp}$ values (e.g., `G5e-05_PR0.1_PF50_PA0.05_PAF25`, $I_{ov} = 0.6445$, $I_{fsp} = 0.6305$, RMSE = 0.8034). This indicates that moderate $I_{ov}$ and $I_{fsp}$ values may be more beneficial to the model's overall performance.

Therefore, when tuning the model, it is crucial not to focus solely on minimizing $I_{ov}$ and $I_{fsp}$ values. Instead, one should take into account RMSE performance and look for parameter combinations that strike the best balance between predictive accuracy and interpretability.

*Discussion and Recommendations.* The parameter sensitivity analysis reveals that the performance of the ADAR-ANFIS model is highly influenced by the selection of hyperparameters, especially the growth threshold, prune rule threshold, and prune attribute parameters. These parameters not only affect the model's predictive accuracy (RMSE) but also play a crucial role in determining the overlap between fuzzy sets ($I_{ov}$) and the precision of fuzzy set positioning ($I_{fsp}$).

A lower growth threshold enhances predictive performance but lead to higher $I_{ov}$ and $I_{fsp}$ values, which could negatively affect the model's discriminability and interpretability. Increasing the prune rule and attribute thresholds appropriately can help reduce $I_{ov}$ and $I_{fsp}$ values while maintaining or even improving the model's predictive performance.

The interaction between parameters plays a crucial role in determining model performance. Optimal performance is not achieved by extreme values of a single parameter but rather by balanced combinations that account for their collective influence. These combinations impact the model's structure and learning dynamics, requiring a balance between model complexity, predictive performance, and structural features.

Based on the experimental findings, the following configuration recommendations for the ADAR-ANFIS model are proposed:

- **Select a moderate growth threshold** (e.g., $5 \times 10^{-5}$) to ensure the model generates enough rules to capture data patterns while preventing excessively high $I_{ov}$ and $I_{fsp}$ values.
- **Appropriately increase the prune rule threshold** (e.g., 0.1) to effectively eliminate less significant rules, reduce $I_{ov}$ values, and enhance the discriminability of the rule base.
- **Set reasonable prune rule and attribute frequencies** (e.g., prune rule frequency of 50 epochs, prune attribute frequency of 25 to 50 epochs) to enable timely pruning, reduce $I_{fsp}$ values, and ensure accurate fuzzy set positioning.
- **Moderately increase the prune attribute threshold** (e.g., 0.1) to minimize unnecessary attributes while preserving essential features, improving the model's interpretability and generalization ability.
- **Continuously monitor $I_{ov}$ and $I_{fsp}$ indicators** to ensure they remain within suitable levels, maintaining an optimal balance between predictive performance and model structure.

## 5 Conclusion

This paper presents the ADAR framework as a solution to the challenges of high-dimensional data in neuro-fuzzy inference systems. By incorporating a dual importance weighting mechanism for rules and attributes, along with adaptive strategies for rule growth and pruning, ADAR achieves substantial improvements in both predictive accuracy and interpretability across multiple benchmark datasets. Experimental results reveal that on datasets with varying size, ranging from 7 to 27, the ADAR-based architecture consistently outperforms traditional and modern fuzzy models, achieving RMSE reductions of up to 9% in tasks such as Boston Housing Price prediction. Additionally, it significantly reduces reliance on large-scale rule bases.

Ablation studies further support the effectiveness of ADAR's key components—dynamic rule and attribute management—and demonstrate how these strategies reduce rule overlap and optimize fuzzy set positioning, resulting in a more compact yet expressive model. These findings show that ADAR not only provides an effective solution for optimizing performance and enhancing transparency in complex, heterogeneous data distributions but also offers a versatile framework for designing fuzzy systems capable of adapting to the varied demands and constraints of different domains.

Looking ahead, the adaptability and efficiency of ADAR open promising directions for real-world applications, particularly in autonomous driving. Future research could explore its integration into real-time decision-making systems for autonomous vehicles, focusing on tasks like adaptive cruise control, lane merging, and collision avoidance. By leveraging ADAR's ability to dynamically rules and attributes, such systems could enhance robustness in handling complex traffic scenarios characterized by high-dimensional and heterogeneous data streams. Furthermore, extending the framework to accommodate multi-agent interactions in autonomous driving would be a critical step toward advancing cooperative and explainable AI systems. This line of research could accelerate the creation of interpretable intelligent systems capable of making safe and efficient decisions in diverse and uncertain driving conditions.

With its proven scalability and robustness, ADAR paves the way for the advancement of interpretable neuro-fuzzy systems in high-dimensional data settings, offering transformative potential across domains such as real-time decision support systems, industrial automation, and intelligent prediction in complex environments. This work not only contributes to the theoretical development of fuzzy inference systems but also lays a strong foundation for their practical deployment in challenging and safety-critical applications.